\documentclass[11pt]{article}
\usepackage{acl2016}
\usepackage{times}
\usepackage{url}
\usepackage{latexsym}
\usepackage{amsfonts}
\usepackage{bm}
\usepackage{graphicx}
\usepackage{algorithm}
\usepackage{algpseudocode}
\usepackage{stmaryrd}
\usepackage{multirow}
\usepackage{color}

\aclfinalcopy 
\def\aclpaperid{***} 

\def\argmax{\mathop{\rm argmax}}%

\title{Semi-Supervised Learning for Neural Machine Translation}

\author{Yong Cheng$^\#$,  Wei Xu$^\#$, Zhongjun He$^+$, Wei He$^+$, Hua Wu$^+$, Maosong Sun$^\dagger$ and Yang Liu$^\dagger$ \thanks{\quad Yang Liu is the corresponding author.}\\
 $^\#$Institute for Interdisciplinary Information Sciences, Tsinghua University, Beijing, China  \\
 $^\dagger$State Key Laboratory of Intelligent Technology and Systems  \\
  Tsinghua National Laboratory for Information Science and Technology \\
 Department of Computer Science and Technology, Tsinghua University, Beijing, China\\
 $^+$Baidu Inc., Beijing, China  \\
 {\tt chengyong3001@gmail.com }
 {\tt weixu@tsinghua.edu.cn } \\
 {\tt \{hezhongjun,hewei06,wu\_hua\}@baidu.com}   \\
 {\tt \{sms,liuyang2011\}@tsinghua.edu.cn }
}

\date{}

\begin{document}
\maketitle

\begin{abstract}
  While end-to-end neural machine translation (NMT) has made remarkable progress recently, NMT systems only rely on parallel corpora for parameter estimation. Since parallel corpora are usually limited in quantity, quality, and coverage, especially for low-resource languages, it is appealing to exploit monolingual corpora to improve NMT. We propose a semi-supervised approach for training NMT models on the concatenation of labeled (parallel corpora) and unlabeled (monolingual corpora) data. The central idea is to reconstruct the monolingual corpora using an autoencoder, in which the source-to-target and target-to-source translation models serve as the encoder and decoder, respectively. Our approach can not only exploit the monolingual corpora of the target language, but also of the source language. Experiments on the Chinese-English dataset show that our approach achieves significant improvements over state-of-the-art SMT and NMT systems.
\end{abstract}

\section{Introduction}
End-to-end neural machine translation (NMT), which leverages a single, large neural network to directly transform a source-language sentence into a target-language sentence, has attracted increasing attention in recent several years \cite{Kalchbrenner:13,Sutskever:14,Bahdanau:15}. Free of latent structure design and feature engineering that are critical in conventional statistical machine translation (SMT) \cite{Brown:93,Koehn:03,Chiang:05}, NMT has proven to excel in modeling long-distance dependencies by enhancing recurrent neural networks (RNNs) with the gating \cite{Hochreiter:97,Cho:14,Sutskever:14} and attention mechanisms \cite{Bahdanau:15}.

However, most existing NMT approaches suffer from a major drawback: they heavily rely on parallel corpora for training translation models. This is because NMT directly models the probability of a target-language sentence given a source-language sentence and does not have a separate language model like SMT \cite{Kalchbrenner:13,Sutskever:14,Bahdanau:15}. Unfortunately, parallel corpora are usually only available for a handful of resource-rich languages and restricted to limited domains such as government documents and news reports. In contrast, SMT is capable of exploiting abundant target-side monolingual corpora to boost fluency of translations. Therefore, the unavailability of large-scale, high-quality, and wide-coverage parallel corpora hinders the applicability of NMT.

As a result, several authors have tried to use abundant monolingual corpora to improve NMT. \newcite{Gulcehre:15} propose two methods, which are referred to as shallow fusion and deep fusion, to integrate a language model into NMT. The basic idea is to use the language model to score the candidate words proposed by the translation model at each time step or concatenating the hidden states of the language model and the decoder. Although their approach leads to significant improvements, one possible downside is that the network architecture has to be modified to integrate the language model.

\begin{figure*}[!t]
\centering
\includegraphics[width=0.8\textwidth]{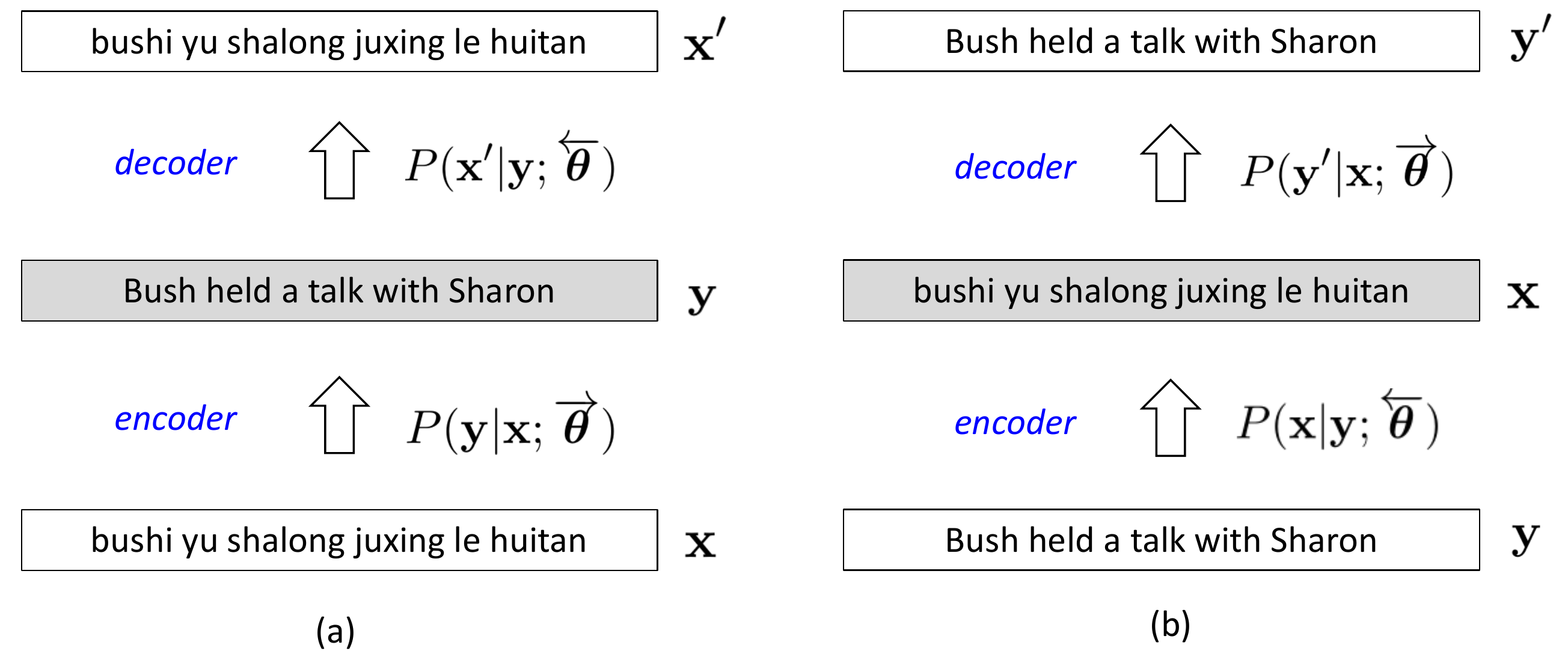}
\caption{Examples of (a) source autoencoder and (b) target autoencoder on monolingual corpora. Our idea is to leverage autoencoders to exploit monolingual corpora for NMT. In a source autoencoder, the source-to-target model $P(\mathbf{y}|\mathbf{x};\overrightarrow{\bm{\theta}})$ serves as an encoder to transform the observed source sentence $\mathbf{x}$ into a latent target sentence $\mathbf{y}$ (highlighted in grey), from which the target-to-source model $P(\mathbf{x}'|\mathbf{y}; \overleftarrow{\bm{\theta}})$ reconstructs a copy of the observed source sentence $\mathbf{x}'$ from the latent target sentence. As a result, monolingual corpora can be combined with parallel corpora to train bidirectional NMT models in a semi-supervised setting.} \label{fig_example}
\end{figure*}

Alternatively, \newcite{Sennrich:15} propose two approaches to exploiting monolingual corpora that is transparent to network architectures.  The first approach pairs monolingual sentences with dummy input. Then, the parameters of encoder and attention model are fixed when training on these pseudo parallel sentence pairs.  In the second approach, they first train a nerual translation model on the parallel corpus and then use the learned model to translate a monolingual corpus. The monolingual corpus and its translations constitute an additional pseudo parallel corpus. Similar ideas have also been suggested in conventional SMT \cite{Ueffing:07,Bertoldi:09}. \newcite{Sennrich:15} report that their approach significantly improves translation quality across a variety of language pairs.

In this paper, we propose semi-supervised learning for neural machine translation. Given labeled (i.e., parallel corpora) and unlabeled (i.e., monolingual corpora) data, our approach jointly trains source-to-target and target-to-source translation models.  The key idea is to append a reconstruction term to the training objective, which aims to reconstruct the observed monolingual corpora using an autoencoder. In the autoencoder, the source-to-target and target-to-source models serve as the encoder and decoder, respectively. As the inference is intractable, we propose to sample the full search space to improve the efficiency. Specifically, our approach has the following advantages:


\begin{enumerate}
\item {\em Transparent to network architectures}: our approach does not depend on specific architectures and can be easily applied to arbitrary end-to-end NMT systems.
\item {\em Both the source and target monolingual corpora can be used}: our approach can benefit NMT not only using target monolingual corpora in a conventional way, but also the monolingual corpora of the source language.
\end{enumerate}

Experiments on Chinese-English NIST datasets show that our approach results in significant improvements in both directions over state-of-the-art SMT and NMT systems.

\section{Semi-Supervised Learning for Neural Machine Translation}
\subsection{Supervised Learning}
Given a parallel corpus $\mathcal{D} = \{ \langle \mathbf{x}^{(n)}, \mathbf{y}^{(n)} \rangle \}_{n=1}^{N}$, the standard training objective in NMT is to maximize the likelihood of the training data:
\begin{eqnarray}
L(\bm{\theta}) = \sum_{n=1}^{N} \log P(\mathbf{y}^{(n)}|\mathbf{x}^{(n)}; \bm{\theta}),
\end{eqnarray}
where $P(\mathbf{y}|\mathbf{x}; \bm{\theta})$ is a neural translation model and $\bm{\theta}$ is a set of model parameters. $\mathcal{D}$ can be seen as {\em labeled} data for the task of predicting a target sentence $\mathbf{y}$ given a source sentence $\mathbf{x}$.

As $P(\mathbf{y}|\mathbf{x}; \bm{\theta})$ is modeled by a single, large neural network, there does not exist a separate target language model $P(\mathbf{y}; \bm{\theta})$ in NMT. Therefore, parallel corpora have been the only resource for parameter estimation in most existing NMT systems. Unfortunately, even for a handful of resource-rich languages, the available domains are unbalanced and restricted to government documents and news reports. Therefore, the availability of large-scale, high-quality, and wide-coverage parallel corpora becomes a major obstacle for NMT.

\subsection{Autoencoders on Monolingual Corpora}

It is appealing to explore the more readily available, abundant monolingual corpora to improve NMT. Let us first consider an {\em unsupervised} setting: how to train NMT models on a monolingual corpus $\mathcal{T}=\{\mathbf{y}^{(t)}\}_{t=1}^{T}$?

Our idea is to leverage {\em autoencoders} \cite{Vincent:10,Socher:11}: (1) {\em encoding} an observed target sentence into a latent source sentence using a target-to-source translation model and (2) {\em decoding} the source sentence to reconstruct the observed target sentence using a source-to-target model. For example, as shown in Figure \ref{fig_example}(b), given an observed English sentence ``Bush held a talk with Sharon'', a target-to-source translation model (i.e., encoder) transforms it into a Chinese translation ``bushi yu shalong juxing le huitan'' that is unobserved on the training data (highlighted in grey). Then, a source-to-target translation model (i.e., decoder) reconstructs the observed English sentence from the Chinese translation.

More formally, let $P(\mathbf{y}|\mathbf{x}; \overrightarrow{\bm{\theta})}$ and $P(\mathbf{x}|\mathbf{y}; \overleftarrow{\bm{\theta}})$ be {\em source-to-target} and {\em target-to-source} translation models respectively, where $\overrightarrow{\bm{\theta}}$ and $\overleftarrow{\bm{\theta}}$ are corresponding model parameters. An autoencoder aims to reconstruct the observed target sentence via a latent source sentence:
\begin{eqnarray}
&&P(\mathbf{y}'|\mathbf{y}; \overrightarrow{\bm{\theta}}, \overleftarrow{\bm{\theta}}) \nonumber \\
&=& \sum_{\mathbf{x}} P(\mathbf{y}', \mathbf{x}|\mathbf{y}; \overrightarrow{\bm{\theta}}, \overleftarrow{\bm{\theta}}) \nonumber \\
&=& \sum_{\mathbf{x}} \underbrace{ P(\mathbf{x}|\mathbf{y}; \overleftarrow{\bm{\theta}})}_{\textrm{{\em encoder}}} \underbrace{ P(\mathbf{y}'|\mathbf{x}; \overrightarrow{\bm{\theta}})}_{\textrm{{\em decoder}}}, \label{eq:autoencoder}
\end{eqnarray}
where $\mathbf{y}$ is an observed target sentence, $\mathbf{y}'$ is a copy of $\mathbf{y}$ to be reconstructed, and $\mathbf{x}$ is a latent source sentence.

We refer to Eq. (\ref{eq:autoencoder}) as a {\em target autoencoder}. \footnote{Our definition of auotoencoders is inspired by Ammar et al. \shortcite{Ammar:14}. Note that our autoencoders inherit the same spirit from conventional autoencoders \cite{Vincent:10,Socher:11} except that the hidden layer is denoted by a latent sentence instead of real-valued vectors.} Likewise, given a monolingual corpus of source language $\mathcal{S}=\{\mathbf{x}^{(s)}\}_{s=1}^{S}$, it is natural to introduce a {\em source autoencoder} that aims at reconstructing the observed source sentence via a latent target sentence:
\begin{eqnarray}
&& P(\mathbf{x}'|\mathbf{x}; \overrightarrow{\bm{\theta}}, \overleftarrow{\bm{\theta}}) \nonumber \\
&=& \sum_{\mathbf{y}} P(\mathbf{x}', \mathbf{y}|\mathbf{x}; \overleftarrow{\bm{\theta}}) \nonumber \\
&=& \sum_{\mathbf{y}} \underbrace{P(\mathbf{y}|\mathbf{x}; \overrightarrow{\bm{\theta}})}_{\textrm{{\em encoder}}} \underbrace{P(\mathbf{x}'|\mathbf{y}; \overleftarrow{\bm{\theta}})}_{\textrm{{\em decoder}}}.
\end{eqnarray}

Please see Figure \ref{fig_example}(a) for illustration.

\subsection{Semi-Supervised Learning}

As the autoencoders involve both source-to-target and target-to-source models, it is natural to combine parallel corpora and monolingual corpora to learn birectional NMT translation models in a semi-supervised setting.

Formally, given a parallel corpus $\mathcal{D} = \{ \langle \mathbf{x}^{(n)}, \mathbf{y}^{(n)} \rangle \}_{n=1}^{N}$ , a monolingual corpus of target language $\mathcal{T} = \{ \mathbf{y}^{(t)} \}_{t=1}^{T}$, and  a monolingual corpus of source language $\mathcal{S} = \{ \mathbf{x}^{(s)} \}_{s=1}^{S}$,  we introduce our new semi-supervised training objective as follows:
\begin{eqnarray}
&&J(\overrightarrow{\bm{\theta}}, \overleftarrow{\bm{\theta}}) \nonumber \\
&=& \underbrace{\sum_{n=1}^{N} \log P(\mathbf{y}^{(n)}|\mathbf{x}^{(n)}; \overrightarrow{\bm{\theta}})}_{\textrm{{\em source-to-target likelihood}}}  \nonumber \\
&& + \underbrace{\sum_{n=1}^{N} \log P(\mathbf{x}^{(n)}|\mathbf{y}^{(n)}; \overleftarrow{\bm{\theta}})}_{\textrm{{\em target-to-source likelihood}}}  \nonumber \\
&& + \lambda_{1} \underbrace{\sum_{t=1}^{T} \log P(\mathbf{y}'|\mathbf{y}^{(t)}; \overrightarrow{\bm{\theta}}, \overleftarrow{\bm{\theta}})}_{\textrm{{\em target autoencoder}}} \nonumber \\
&& + \lambda_{2} \underbrace{\sum_{s=1}^{S} \log P(\mathbf{x}'|\mathbf{x}^{(s)}; \overrightarrow{\bm{\theta}}, \overleftarrow{\bm{\theta}})}_{\textrm{{\em source autoencoder}}}, \label{eq:objective}
\end{eqnarray}
where $\lambda_{1}$ and $\lambda_{2}$ are hyper-parameters for balancing the preference between likelihood and autoencoders.

Note that the objective consists of four parts: source-to-target likelihood, target-to-source likelihood, target autoencoder, and source autoencoder. In this way, our approach is capable of exploiting abundant monolingual corpora of both source and target languages.

The optimal model parameters are given by
\begin{eqnarray}
\overrightarrow{\bm{\theta}}^{*} = \argmax\Bigg\{ \sum_{n=1}^{N}\log P(\mathbf{y}^{(n)}|\mathbf{x}^{(n)}; \overrightarrow{\bm{\theta}}) + \nonumber \\
\lambda_{1} \sum_{t=1}^{T}\log P(\mathbf{y}'|\mathbf{y}^{(t)}; \overrightarrow{\bm{\theta}}, \overleftarrow{\bm{\theta}}) + \nonumber \\
\lambda_{2} \sum_{s=1}^{S}\log P(\mathbf{x}'|\mathbf{x}^{(s)}; \overrightarrow{\bm{\theta}}, \overleftarrow{\bm{\theta}})   \Bigg\} \\
\overleftarrow{\bm{\theta}}^{*} = \argmax\Bigg\{ \sum_{n=1}^{N}\log P(\mathbf{x}^{(n)}|\mathbf{y}^{(n)}; \overleftarrow{\bm{\theta}}) + \nonumber \\
\lambda_{1} \sum_{t=1}^{T}\log P(\mathbf{y}'|\mathbf{y}^{(t)}; \overrightarrow{\bm{\theta}}, \overleftarrow{\bm{\theta}}) + \nonumber \\
\lambda_{2} \sum_{s=1}^{S}\log P(\mathbf{x}'|\mathbf{x}^{(s)}; \overrightarrow{\bm{\theta}}, \overleftarrow{\bm{\theta}})   \Bigg\}
\end{eqnarray}

It is clear that the source-to-target and target-to-source models are connected via the autoencoder and can hopefully benefit each other in joint training.

\subsection{Training}

We use mini-batch stochastic gradient descent to train our joint model. For each iteration, besides the mini-batch from the parallel corpus, we also construct two additional mini-batches by randomly selecting sentences from the source and target monolingual corpora. Then, gradients are collected from these mini-batches to update model parameters.

The partial derivative of $J(\overrightarrow{\bm{\theta}}, \overleftarrow{\bm{\theta}})$ with respect to the source-to-target model $\overrightarrow{\bm{\theta}}$ is given by

\begin{eqnarray}
&& \frac{\partial J(\overrightarrow{\bm{\theta}}, \overleftarrow{\bm{\theta}})}{\partial \overrightarrow{\bm{\theta}}}   \nonumber \\
&=& \sum_{n=1}^{N} \frac{\partial \log P(\mathbf{y}^{(n)}|\mathbf{x}^{(n)}; \overrightarrow{\bm{\theta}})}{\partial \overrightarrow{\bm{\theta}}}  \nonumber \\
&& + \lambda_{1} \sum_{t=1}^{T} \frac{\partial\log P(\mathbf{y}'|\mathbf{y}^{(t)}; \overrightarrow{\bm{\theta}}, \overleftarrow{\bm{\theta}})}{\partial \overrightarrow{\bm{\theta}}} \nonumber \\
&& +\lambda_{2} \sum_{s=1}^{S} \frac{\partial\log P(\mathbf{x}'|\mathbf{x}^{(s)}; \overrightarrow{\bm{\theta}}, \overleftarrow{\bm{\theta}})}{\partial \overrightarrow{\bm{\theta}}}. \label{eq:derivative}
\end{eqnarray}
The partial derivative with respect to $\overleftarrow{\bm{\theta}}$ can be calculated similarly.

Unfortunately, the second and third terms in Eq. (\ref{eq:derivative}) are intractable to calculate due to the exponential search space. For example, the derivative in the third term in Eq. (\ref{eq:derivative}) is given by
\begin{eqnarray}
\frac{\sum_{\mathbf{x} \in \mathcal{X}(\mathbf{y})}P(\mathbf{x}|\mathbf{y}; \overleftarrow{\bm{\theta}})P(\mathbf{y}'|\mathbf{x}; \overrightarrow{\bm{\theta}})\frac{\partial \log P(\mathbf{y}'|\mathbf{x}; {\tiny \overrightarrow{\bm{\theta}}})}{\partial {\tiny \overrightarrow{\bm{\theta}}}}}{\sum_{\mathbf{x} \in \mathcal{X}(\mathbf{y})}P(\mathbf{x}|\mathbf{y}; \overleftarrow{\bm{\theta}})P(\mathbf{y}'|\mathbf{x}; \overrightarrow{\bm{\theta}})}.
\label{eq:intractable}
\end{eqnarray}
It is prohibitively expensive to compute the sums due to the exponential search space of $\mathcal{X}(\mathbf{y})$.

Alternatively, we propose to use a subset of the full space $\tilde{\mathcal{X}}(\mathbf{y}) \subset \mathcal{X}(\mathbf{y})$ to approximate Eq. (\ref{eq:intractable}):

\begin{eqnarray}
\frac{\sum_{\mathbf{x} \in \tilde{\mathcal{X}}(\mathbf{y})}P(\mathbf{x}|\mathbf{y}; \overleftarrow{\bm{\theta}})P(\mathbf{y}'|\mathbf{x}; \overrightarrow{\bm{\theta}})\frac{\partial \log P(\mathbf{y}'|\mathbf{x}; {\tiny \overrightarrow{\bm{\theta}}})}{\partial {\tiny \overrightarrow{\bm{\theta}}}}}{\sum_{\mathbf{x} \in \tilde{\mathcal{X}}(\mathbf{y})}P(\mathbf{x}|\mathbf{y}; \overleftarrow{\bm{\theta}})P(\mathbf{y}'|\mathbf{x}; \overrightarrow{\bm{\theta}})}. \label{eq:approx}
\end{eqnarray}

In practice, we use the top-$k$ list of candidate translations of $\mathbf{y}$ as $\tilde{\mathcal{X}}(\mathbf{y})$. As $|\tilde{\mathcal{X}}(\mathbf{y})| \ll \mathcal{X}|(\mathbf{y})|$, it is possible to calculate Eq. (\ref{eq:approx}) efficiently by enumerating all candidates in $\tilde{\mathcal{X}}(\mathbf{y})$. In practice, we find this approximation results in significant improvements and $k=10$ seems to suffice to keep the balance between efficiency and translation quality. 

\section{Experiments}

\subsection{Setup}

\begin{table}[!t]
\centering
\begin{tabular}{|l|l||r|r|}
\hline
& & Chinese & English \\
\hline \hline
 & \# Sent. & \multicolumn{2}{c|}{2.56M} \\
 \cline{2-4}
Parallel & \# Word & 67.54M & 74.82M \\
 \cline{2-4}
 & Vocab. & 0.21M & 0.16M \\
 \hline \hline
 & \# Sent. & 18.75M & 22.32M \\
 \cline{2-4}
Monolingual & \# Word & 451.94M & 399.83M \\ \cline{2-4}
 & Vocab. & 0.97M & 1.34M \\
 \hline
\end{tabular}
\caption{Characteristics of parallel and monolingual corpora.} \label{table:data}
\end{table}

We evaluated our approach on the Chinese-English dataset.

\begin{figure}[!t]
\centering
\includegraphics[width=0.5\textwidth]{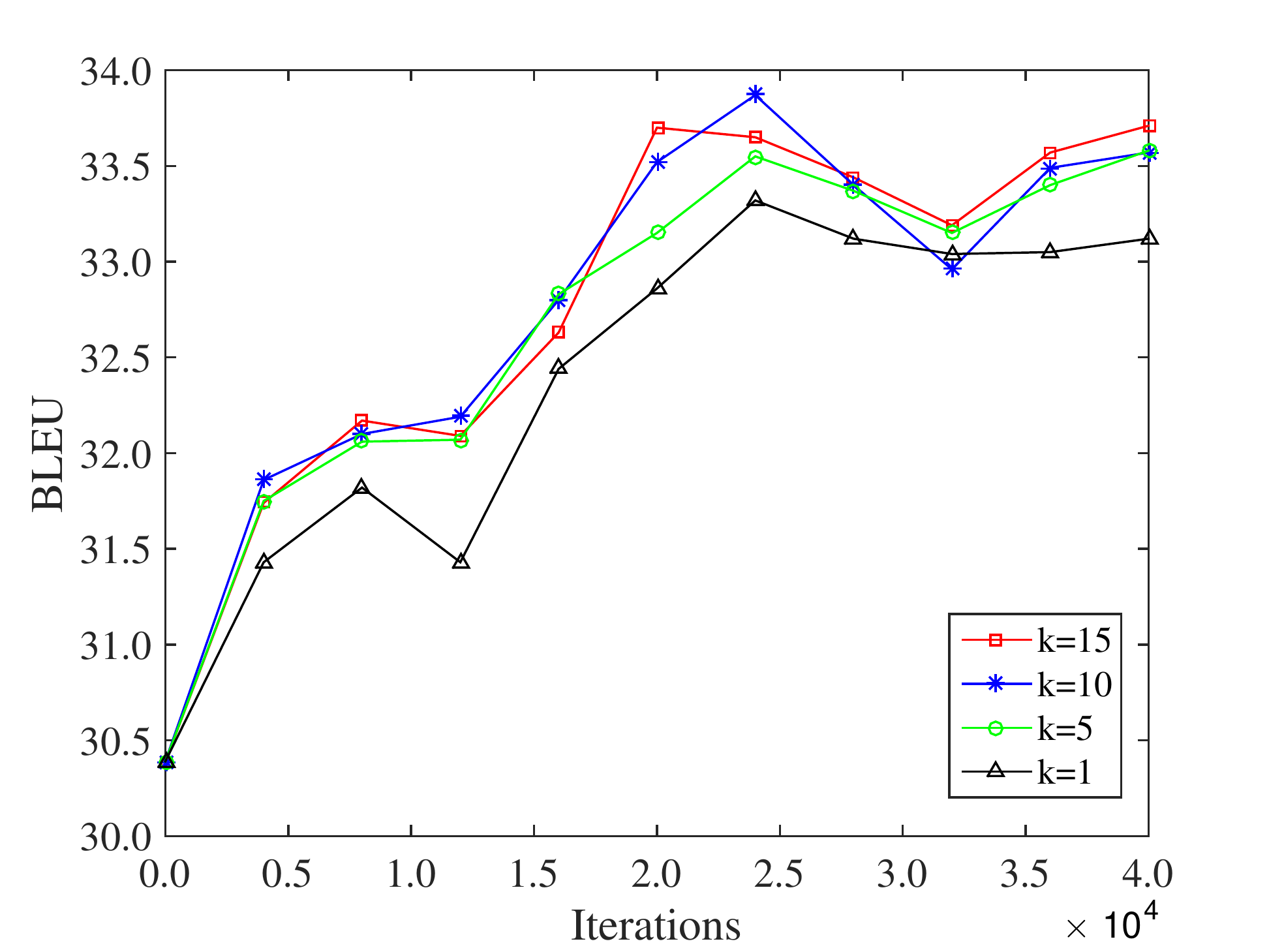}
\caption{Effect of sample size $k$ on the Chinese-to-English validation set.} \label{fig:sample_size_ce}
\end{figure}

\begin{figure}[!t]
\centering
\includegraphics[width=0.5\textwidth]{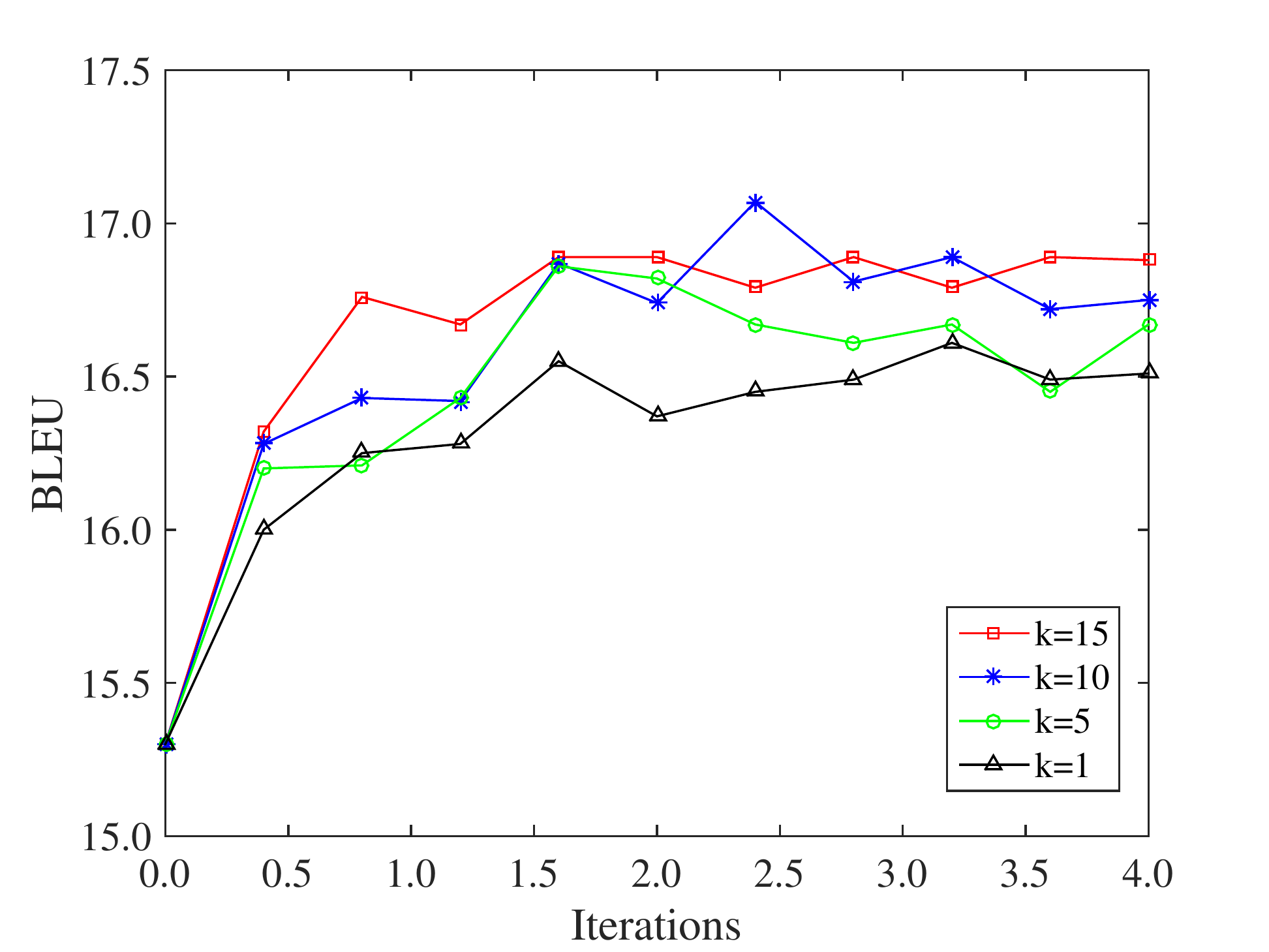}
\caption{Effect of sample size $k$ on the English-to-Chinese validation set.} \label{fig:sample_size_ec}
\end{figure}

As shown in Table \ref{table:data}, we use both a parallel corpus and two monolingual corpora as the training set. The parallel corpus from LDC consists of 2.56M sentence pairs with 67.53M Chinese words and 74.81M English words. The vocabulary sizes of Chinese and English are 0.21M and 0.16M, respectively. We use the Chinese and English parts of the Xinhua portion of the GIGAWORD corpus as the monolingual corpora. The Chinese monolingual corpus contains 18.75M sentences with 451.94M words. The English corpus contains 22.32M sentences with 399.83M words. The vocabulary sizes of Chinese and English are 0.97M and 1.34M, respectively.

For Chinese-to-English translation, we use the NIST 2006 Chinese-English dataset as the validation set for hyper-parameter optimization and model selection. The NIST 2002, 2003, 2004, and 2005 datasets serve as test sets. Each Chinese sentence has four reference translations. For English-to-Chinese translation, we use the NIST datasets in a reverse direction: treating the first English sentence in the four reference translations as a source sentence and the original input Chinese sentence as the single reference translation. The evaluation metric is case-insensitive BLEU \cite{Papineni:02} as calculated by the \verb|multi-bleu.perl| script.

We compared our approach with two state-of-the-art SMT and NMT systems:
\begin{enumerate}
\item \textproc{Moses} \cite{Koehn:07}: a phrase-based SMT system;
\item \textproc{RNNsearch} \cite{Bahdanau:15}: an attention-based NMT system.
\end{enumerate}

For \textproc{Moses}, we use the default setting to train the phrase-based translation on the parallel corpus and optimize the parameters of log-linear models using the minimum error rate training algorithm \cite{Och:03}. We use the SRILM toolkit \cite{Stolcke:02} to train 4-gram language models.

\begin{figure}[!t]
\centering
\includegraphics[width=0.5\textwidth]{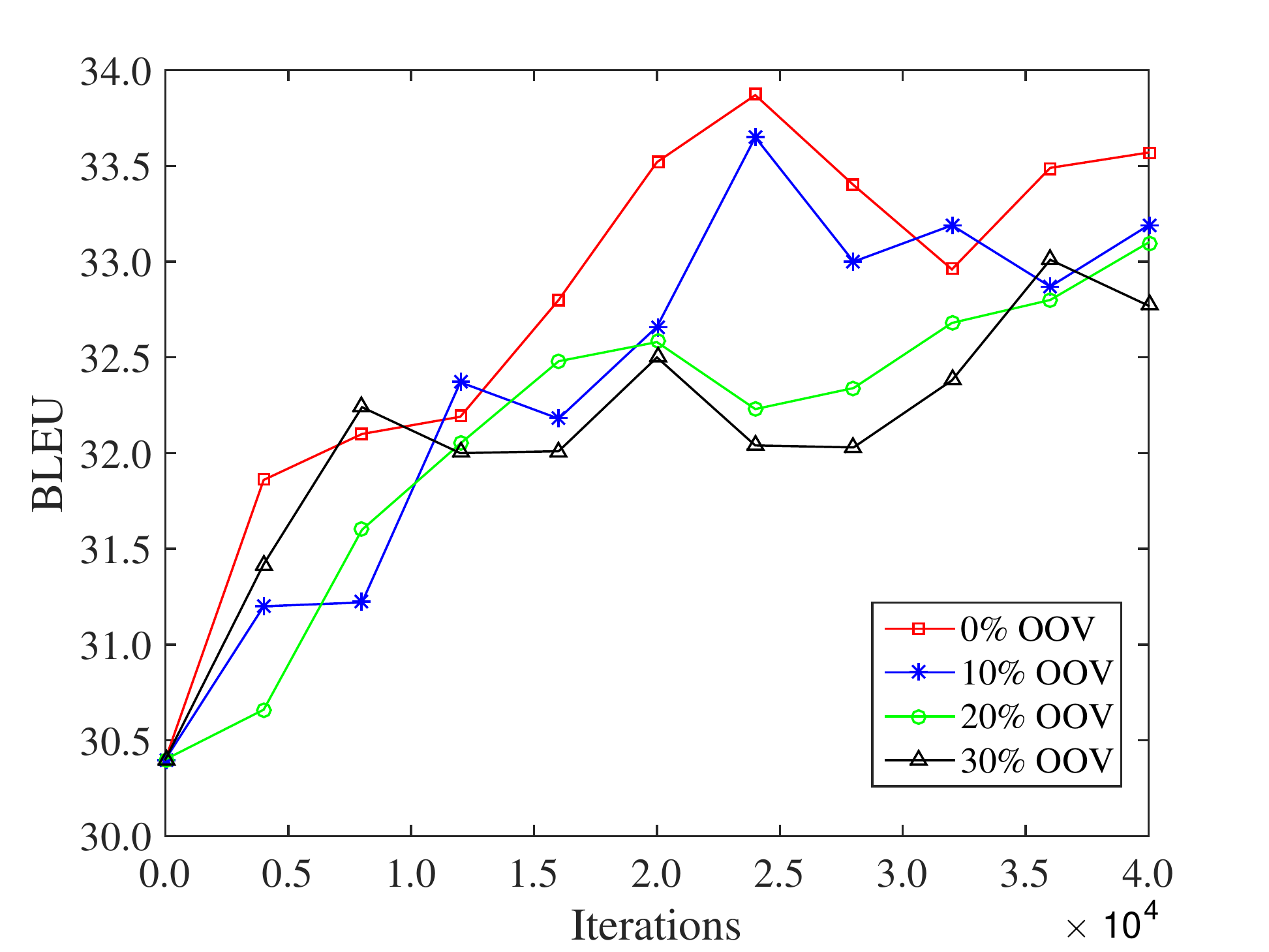}
\caption{Effect of OOV ratio on the Chinese-to-English validation set.} \label{fig:oov_ce}
\end{figure}

\begin{figure}[!t]
\centering
\includegraphics[width=0.5\textwidth]{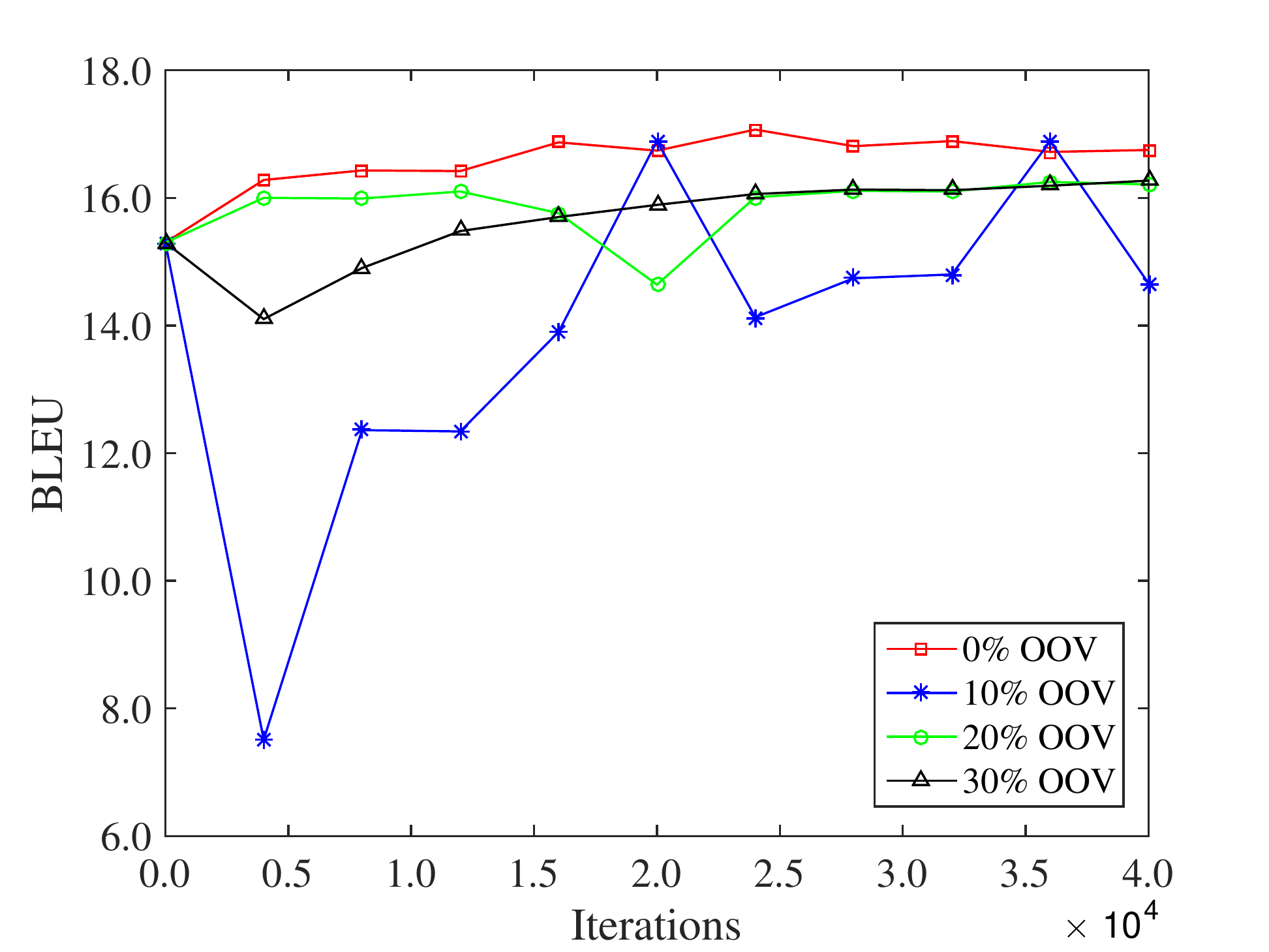}
\caption{Effect of OOV ratio on the English-to-Chinese validation set.} \label{fig:oov_ec}
\end{figure}

For \textproc{RNNSearch}, we use the parallel corpus to train the attention-based neural translation models. We set the vocabulary size of word embeddings to 30K for both Chinese and English. We follow \newcite{Luong:15} to address rare words.

On top of \textproc{RNNSearch}, our approach is capable of training bidirectional attention-based neural translation models on the concatenation of parallel and monolingual corpora. The sample size $k$ is set to 10. We set the hyper-parameter $\lambda_{1}=0.1$ and $\lambda_{2}=0$ when we add the target monolingual corpus, and $\lambda_{1}=0$ and $\lambda_{2}=0.1$ for source monolingual corpus incorporation. The threshold of gradient clipping is set to $0.05$. The parameters of our model are initialized by the  model trained on parallel corpus.

\subsection{Effect of Sample Size $k$}

As the inference of our approach is intractable, we propose to approximate the full search space with the top-$k$ list of candidate translations to improve efficiency (see Eq. (\ref{eq:approx})).

Figure \ref{fig:sample_size_ce} shows the BLEU scores of various settings of $k$ over time. Only the English monolingual corpus is appended to the training data. We observe that increasing the size of the approximate search space generally leads to improved BLEU scores. There are significant gaps between $k=1$ and $k=5$. However, keeping increasing $k$ does not result in significant improvements and decreases the training efficiency. We find that $k=10$ achieves a balance between training efficiency and translation quality. As shown in Figure \ref{fig:sample_size_ec}, similar findings are also observed on the English-to-Chinese validation set. Therefore, we set $k=10$ in the following experiments.

\subsection{Effect of OOV Ratio}

\begin{table*}[!t]
\small
\centering
\begin{tabular}{|c|c|c|c|c||l|l|l|l|l|}
\hline
\multirow{2}{*}{System} & \multicolumn{3}{|c|}{Training Data} & \multirow{2}{*}{Direction} & \multirow{2}{*}{NIST06} & \multirow{2}{*} {NIST02} & \multirow{2}{*}{NIST03} & \multirow{2}{*}{NIST04} & \multirow{2}{*}{NIST05}  \\
\cline{2-4}
&CE &C & E & &  & & & &  \\
\hline \hline
\multirow{4}{*}{\textproc{Moses}} & \multirow{2}{*}{$\surd$}  & \multirow{2}{*}{$\times$} & \multirow{2}{*}{$\times$} & C $\rightarrow$ E &32.48 &32.69 &32.39 &33.62 &30.23  \\
& & &  & E $\rightarrow$ C &14.27 &18.28  &15.36 &13.96 &14.11 \\
\cline{2-10}
  &$\surd$ &$\times$ & $\surd$  & C $\rightarrow$ E  &34.59 &35.21  &35.71 &35.56 &33.74   \\
 &$\surd$ &$\surd$ & $\times$  &  E $\rightarrow$ C &20.69 & 25.85 &19.76 &18.77 & 19.74  \\
\hline \hline

\multirow{6}{*}{\textproc{RNNsearch}} & \multirow{2}{*}{$\surd$}  & \multirow{2}{*}{$\times$} & \multirow{2}{*}{$\times$} & C $\rightarrow$E &30.74 & 35.16 &33.75 &34.63 &31.74 \\
& & & &  E $\rightarrow$C &15.71 &20.76  &16.56 &16.85 &15.14  \\

\cline{2-10}

 & \multirow{2}{*}{$\surd$}  & \multirow{2}{*}{$\times$} & \multirow{2}{*}{$\surd$} & C $\rightarrow$ E &35.61$^{**++}$ &38.78$^{**++}$  &38.32$^{**++}$ &38.49$^{**++}$ &36.45$^{**++}$  \\
& & & &  E $\rightarrow$C &17.59$^{++}$ &23.99 $^{++}$  &18.95$^{++}$ &18.85$^{++}$ &17.91$^{++}$  \\
\cline{2-10}

& \multirow{2}{*}{$\surd$}  & \multirow{2}{*}{$\surd$} & \multirow{2}{*}{$\times$} & C $\rightarrow$E & 35.01$^{++}$ &38.20$^{**++}$  &37.99$^{**++}$ &38.16$^{**++}$ &36.07$^{**++}$  \\
& & & &  E $\rightarrow$C &21.12$^{*++}$ &29.52$^{**++}$  &20.49$^{**++}$ &21.59$^{**++}$ &19.97$^{++}$ \\
\hline
\end{tabular}
\caption{Comparison with \textproc{Moses} and \textproc{RNNsearch}. \textproc{Moses} is a phrase-based statistical machine translation system \cite{Koehn:07}. \textproc{RNNsearch} is an attention-based neural machine translation system \cite{Bahdanau:15}. ``CE'' donates Chinese-English parallel corpus, ``C'' donates Chinese monolingual corpus, and ``E'' donates English monolingual corpus. ``$\surd$'' means the corpus is included in the training data and $\times$ means not included. ``NIST06'' is the validation set and ``NIST02-05'' are test sets. The BLEU scores are case-insensitive. ``*'': significantly better than MOSES ($p<0.05$); ``**'': significantly better than MOSES ($p<0.01$);``+'': significantly better than \textproc{RNNsearch} ($p<0.05$); ``++'': significantly better than \textproc{RNNsearch} ($p<0.01$).}
\label{table:comparison_moses}
\end{table*}

\begin{table*}[!t]
\small
\centering
\begin{tabular}{|c|c|c|c|c|l|l|l|l|l|}
\hline
\multirow{2}{*}{Method} & \multicolumn{3}{|c|}{Training Data} & \multirow{2}{*}{Direction} & \multirow{2}{*}{NIST06} & \multirow{2}{*} {NIST02} & \multirow{2}{*}{NIST03} & \multirow{2}{*}{NIST04} & \multirow{2}{*}{NIST05}  \\
\cline{2-4}
&CE &C &E  & &  & & & &  \\
\hline \hline

\multirow{2}{*}{\newcite{Sennrich:15}} & $\surd$  & $\times$ & $\surd$ & C $\rightarrow$E &34.10 & 36.95 &36.80 &37.99 &35.33 \\
& $\surd$  & $\surd$ & $\times$ &  E $\rightarrow$C &19.85 &28.83  &20.61 &20.54 &19.17  \\
\hline

\multirow{4}{*}{{\em this work}} & \multirow{2}{*}{$\surd$}  & \multirow{2}{*}{$\times$} & \multirow{2}{*}{$\surd$} & C $\rightarrow$E &35.61$^{**}$ &38.78$^{**}$  &38.32$^{**}$ &38.49$^{*}$ &36.45$^{**}$ \\
& & & &  E $\rightarrow$C &17.59 &23.99  &18.95 &18.85 &17.91  \\
\cline{2-10}

& \multirow{2}{*}{$\surd$}  & \multirow{2}{*}{$\surd$} & \multirow{2}{*}{$\times$} & C $\rightarrow$E & 35.01$^{**}$ &38.20$^{**}$  &37.99$^{**}$ &38.16 &36.07$^{**}$ \\
& & & &  E $\rightarrow$C &21.12$^{**}$ &29.52$^{**}$  &20.49 &21.59$^{**}$ &19.97$^{**}$ \\
\hline
\end{tabular}
\caption{Comparison with \newcite{Sennrich:15}. Both \newcite{Sennrich:15} and our approach build on top of \textproc{RNNsearch} to exploit monolingual corpora. The BLEU scores are case-insensitive. ``*'': significantly better than \newcite{Sennrich:15} ($p<0.05$); ``**'': significantly better than \newcite{Sennrich:15} ($p<0.01$).} \label{table:comparison_edin}
\end{table*}

Given a parallel corpus, what kind of monolingual corpus is most beneficial for improving translation quality? To answer this question, we investigate the effect of {\em OOV ratio} on translation quality, which is defined as
\begin{eqnarray}
\textrm{ratio} = \frac{\sum_{y \in \mathbf{y}} \llbracket y \notin \mathcal{V}_{D_t} \rrbracket }{|\mathbf{y}|},
\end{eqnarray}
where $\mathbf{y}$ is a target-language sentence in the monolingual corpus $\mathcal{T}$, $y$ is a target-language word in $\mathbf{y}$, $\mathcal{V}_{D_t}$ is the vocabulary of the target side of the parallel corpus $D$.

Intuitively, the OOV ratio indicates how a sentence in the monolingual resembles the parallel corpus. If the ratio is 0, all words in the monolingual sentence also occur in the parallel corpus.

Figure \ref{fig:oov_ce} shows the effect of OOV ratio on the Chinese-to-English validation set. Only English monolingual corpus is appended to the parallel corpus during training. We constructed four monolingual corpora of the same size in terms of sentence pairs. ``0\% OOV'' means the OOV ratio is 0\% for all sentences in the monolingual corpus. ``10\% OOV'' suggests that the OOV ratio is no greater 10\% for each sentence in the monolingual corpus. We find that using a monolingual corpus with a lower OOV ratio generally leads to higher BLEU scores. One possible reason is that low-OOV monolingual corpus is relatively easier to reconstruct than its high-OOV counterpart and results in better estimation of model parameters.

Figure \ref{fig:oov_ec} shows the effect of OOV ratio on the English-to-Chinese validation set. Only English monolingual corpus is appended to the parallel corpus during training. We find that ``0\% OOV'' still achieves the highest BLEU scores.

\begin{table*}[!t]
\centering
\begin{tabular}{|l|p{1.5\columnwidth}|}
\hline
Monolingual & hongsen shuo , ruguo you na jia famu gongsi dangan yishenshifa , name tamen jiang zihui qiancheng . \\
\hline
Reference &  hongsen said, if any \textcolor{blue}{\textit{logging companies}} dare to defy the law, then they will  \textcolor{blue}{\textit{destroy their own future}} . \\
\hline \hline
Translation &  hun sen said , if any of \textcolor{blue}{\textit{those companies}} dare defy the law , then they will \textcolor{blue}{\textit{have their own fate}} . [{\em iteration 0}] \\
\cline{2-2}
& hun sen said if any \textcolor{blue}{\textit{tree felling company}} dared to break the law , then they would \textcolor{blue}{\textit{kill themselves}} . [{\em iteration 40K}] \\
\cline{2-2}
& hun sen said if any \textcolor{blue}{\textit{logging companies}} dare to defy the law , they would \textcolor{blue}{\textit{destroy the future themselves}} . [{\em iteration 240K}] \\
\hline
\end{tabular}

\begin{tabular}{c}
\\
\end{tabular}

\begin{tabular}{|l|p{1.5\columnwidth}|}
\hline
Monolingual & dan yidan panjue jieguo zuizhong queding , ze bixu zai 30 tian nei zhixing . \\
\hline
Reference & But once \textcolor{blue}{\textit{ the final verdict is confirmed}} , it must be executed within 30 days . \\
\hline \hline
Translation &  however , \textcolor{blue}{\textit{in the final analysis}} , it must be carried out within 30 days . [{\em iteration 0}] \\
\cline{2-2}
& however ,\textcolor{blue}{\textit{ in the final analysis}} , the final decision will be carried out within 30 days . [{\em iteration 40K}] \\
\cline{2-2}
& however , once \textcolor{blue}{\textit{the verdict is finally confirmed}} , it must be carried out within 30 days . [{\em iteration 240K}] \\
\hline
\end{tabular}
\caption{Example translations of sentences in the monolingual corpus during semi-supervised learning. We find our approach is capable of generating better translations of the monolingual corpus over time.} \label{table:example}
\end{table*}

\subsection{Comparison with SMT}

Table \ref{table:comparison_moses} shows the comparison between \textproc{Moses} and our work. \textproc{Moses} used the monolingual corpora as shown in Table \ref{table:data}: 18.75M Chinese sentences and 22.32M English sentences. We find that exploiting monolingual corpora dramatically improves translation performance in both Chinese-to-English and English-to-Chinese directions.

Relying only on parallel corpus, \textproc{RNNsearch} outperforms \textproc{Moses} trained also only on parallel corpus. But the capability of making use of abundant monolingual corpora enables \textproc{Moses} to achieve much higher BLEU scores than \textproc{RNNsearch} only using parallel corpus.

Instead of using all sentences in the monolingual corpora, we constructed smaller monolingual corpora with zero OOV ratio: 2.56M Chinese sentences with 47.51M words and 2.56M English English sentences with 37.47M words. In other words, the monolingual corpora we used in the experiments are much smaller than those used by \textproc{Moses}.

By adding English monolingual corpus, our approach achieves substantial improvements over \textproc{RNNsearch} using only parallel corpus (up to +4.7 BLEU points). In addition, significant improvements are also obtained over \textproc{Moses} using both parallel and monolingual corpora (up to +3.5 BLEU points).

An interesting finding is that adding English monolingual corpora helps to improve English-to-Chinese translation over \textproc{RNNsearch} using only parallel corpus (up to +3.2 BLEU points), suggesting that our approach is capable of improving NMT using source-side monolingual corpora.

In the English-to-Chinese direction, we obtain similar findings. In particular, adding Chinese monolingual corpus leads to more benefits to English-to-Chinese translation than adding English monolingual corpus. We also tried to use both Chinese and English monolingual corpora through simply setting all the $\lambda$ to $0.1$ but failed to obtain further significant improvements.

Therefore, our findings can be summarized as follows:
\begin{enumerate}
\item Adding target monolingual corpus improves over using only parallel corpus for source-to-target translation;
\item Adding source monolingual corpus also improves over using only parallel corpus for source-to-target translation, but the improvements are smaller than adding target monolingual corpus;
\item Adding both source and target monolingual corpora does not lead to further significant improvements.
\end{enumerate}

\subsection{Comparison with Previous Work}
We re-implemented \newcite{Sennrich:15}'s method on top of \textproc{RNNsearch} as follows:
\begin{enumerate}
\item Train the target-to-source neural translation model $P(\mathbf{x}|\mathbf{y}; \overleftarrow{\bm{\theta}})$ on the parallel corpus $D=\{\langle \mathbf{x}^{(n)}, \mathbf{y}^{(n)} \rangle\}_{n=1}^{N}$.
\item The trained target-to-source model $\overleftarrow{\bm{\theta}}^{*}$ is used to translate a target monolingual corpus $\mathcal{T}=\{\mathbf{y}^{(t)}\}_{t=1}^{T}$ into a source monolingual corpus $\tilde{\mathcal{S}}=\{ \tilde{\mathbf{x}}^{(t)}\}_{t=1}^{T}$.
 \item The target monolingual corpus is paired with its translations  to form a pseudo parallel corpus, which is then appended to the original parallel corpus to obtain a larger parallel corpus: $\tilde{\mathcal{D}}=\mathcal{D} \cup \langle \tilde{\mathcal{S}}, \mathcal{T} \rangle$.
 \item Re-train the the source-to-target neural translation model on $\tilde{\mathcal{D}}$ to obtain the final model parameters $\overrightarrow{\bm{\theta}}^{*}$.
\end{enumerate}

Table \ref{table:comparison_edin} shows the comparison results. Both the two approaches use the same parallel and monolingual corpora. Our approach achieves significant improvements over \newcite{Sennrich:15} in both Chinese-to-English and English-to-Chinese directions (up to +1.8 and +1.0 BLEU points). One possible reason is that \newcite{Sennrich:15} only use the pesudo parallel corpus for parameter estimation for once (see Step 4 above) while our approach enables source-to-target and target-to-source models to interact with each other iteratively on both parallel and monolingual corpora.

To some extent, our approach can be seen as an iterative extension of \newcite{Sennrich:15}'s approach: after estimating model parameters on the pseudo parallel corpus, the learned model parameters are used to produce a better pseudo parallel corpus. Table \ref{table:example} shows example Viterbi translations on the Chinese monolingual corpus over iterations:
\begin{eqnarray}
\mathbf{x}^{*} = \argmax_{\mathbf{x}} \Big\{ P(\mathbf{y}'|\mathbf{x}; \overrightarrow{\bm{\theta}}) P(\mathbf{x}|\mathbf{y}; \overleftarrow{\bm{\theta}}) \Big\}.
\end{eqnarray}

We observe that the quality of Viterbi translations generally improves over time.

\section{Related Work}

Our work is inspired by two lines of research: (1) exploiting monolingual corpora for machine translation and (2) autoencoders in unsupervised and semi-supervised learning.

\subsection{Exploiting Monolingual Corpora for Machine Translation}
Exploiting monolingual corpora for conventional SMT has attracted intensive attention in recent years. Several authors have introduced transductive learning to make full use of monolingual corpora \cite{Ueffing:07,Bertoldi:09}. They use an existing translation model to translate unseen source text, which can be paired with its translations to form a pseudo parallel corpus. This process iterates until convergence. While \newcite{Klementiev:12} propose an approach to estimating phrase translation probabilities from monolingual corpora, \newcite{Zhang:13} directly extract parallel phrases from monolingual corpora using retrieval techniques. Another important line of research is to treat translation on monolingual corpora as a decipherment problem \cite{Ravi:11,Dou:14}.

Closely related to \newcite{Gulcehre:15} and \newcite{Sennrich:15}, our approach focuses on learning birectional NMT models via autoencoders on monolingual corpora. The major advantages of our approach are the transparency to network architectures and the capability to exploit both source and target monolingual corpora.

\subsection{Autoencoders in Unsupervised and Semi-Supervised Learning}
Autoencoders and their variants have been widely used in unsupervised deep learning (\cite{Vincent:10,Socher:11,Ammar:14}, just to name a few). Among them, \newcite{Socher:11}'s approach bears close resemblance to our approach as they introduce semi-supervised recursive autoencoders for sentiment analysis. The difference is that we are interested in making a better use of parallel and monolingual corpora while they concentrate on injecting partial supervision to conventional unsupervised autoencoders. \newcite{Dai:15} introduce a sequence autoencoder to reconstruct an observed sequence via RNNs. Our approach differs from sequence autoencoders in that we use bidirectional translation models as encoders and decoders to enable them to interact within the autoencoders.

\section{Conclusion}
We have presented a semi-supervised approach to training bidirectional neural machine translation models. The central idea is to introduce autoencoders on the monolingual corpora with source-to-target and target-to-source translation models as encoders and decoders. Experiments on Chinese-English NIST datasets show that our approach leads to significant improvements.

As our method is sensitive to the OOVs present in monolingual corpora, we plan to integrate \newcite{Jean:15}'s technique on using very large vocabulary into our approach. It is also necessary to further validate the effectiveness of our approach on more language pairs and NMT architectures. Another interesting direction is to enhance the connection between source-to-target and target-to-source models (e.g., letting the two models share the same word embeddings) to help them benefit more from interacting with each other.

\section*{Acknowledgements}

 This work was done while Yong Cheng was visiting Baidu. This research is supported by the 973 Program (2014CB340501, 2014CB340505), the National Natural Science Foundation of China (No. 61522204, 61331013, 61361136003), 1000 Talent Plan grant, Tsinghua Initiative Research Program grants 20151080475 and a Google Faculty Research Award. We sincerely thank the viewers for their valuable suggestions.

\bibliography{acl2016_semi}
\bibliographystyle{acl2016}

\end{document}